%% file: main.tex
\documentclass[10pt,twocolumn,letterpaper]{article}

\usepackage{cvpr}              

\input{preamble}

\definecolor{rowhighlight}{gray}{0.9}
\definecolor{training}{RGB}{103,169,207}
\definecolor{inference}{RGB}{239,138,98}

\definecolor{grey}{RGB}{80,80,80}
\definecolor{darkgrey}{RGB}{55,55,55}
\definecolor{lightgrey}{RGB}{170,170,170}
\definecolor{cvprblue}{rgb}{0.21,0.49,0.74}
\usepackage[pagebackref,breaklinks,colorlinks,citecolor=cvprblue]{hyperref}
\usepackage{times}
\usepackage{epsfig}
\usepackage{graphicx}
\usepackage{amsmath}
\usepackage{amssymb}
\usepackage{comment}

\DeclareMathOperator*{\argmin}{arg\,min}
\newcommand\norm[1]{\lVert#1\rVert}
\usepackage{color}
\usepackage{multirow}
\usepackage{subcaption}
\usepackage{siunitx}
\usepackage{xcolor}
\usepackage{color, colortbl}
\usepackage{cuted}
\usepackage{changepage}
\usepackage{booktabs}
\def\paperID{15421} 
\def\confName{CVPR}
\def\confYear{2024}

\definecolor{shared}{RGB}{103,169,207}
\definecolor{individual}{RGB}{239,138,98}
\definecolor{fixed}{HTML}{6CDAEE}

\title{Neural Parametric Gaussians for Monocular Non-Rigid Object Reconstruction}

\author{Devikalyan Das\textsuperscript{1}
\and
Christopher Wewer\textsuperscript{2}\\
\and
Raza Yunus\textsuperscript{1,2}\\
\and
Eddy Ilg\textsuperscript{1}\\
\and
Jan Eric Lenssen\textsuperscript{2}\\
\and
\textsuperscript{1}Saarland University, Saarland Informatics Campus, Germany\\
\textsuperscript{2}Max Planck Institute for Informatics, Saarland Informatics Campus, Germany\\
{\tt\small \{ddas, jlenssen\}@mpi-inf.mpg.de}
}

\begin{document}
\makeatletter
\let\@oldmaketitle\@maketitle
\renewcommand{\@maketitle}{\@oldmaketitle

\newenvironment{nonfloat}
  {%
   \par\nopagebreak\vspace{\medskipamount}%
   \noindent\begin{minipage}{\linewidth}
   \captionsetup[subfigure]{
     margin=0pt,font+=small,labelformat=parens,labelsep=space,
     skip=-6pt,list=false,hypcap=false
   }%
  }
  {\end{minipage}}

\begin{center}
    \begin{adjustwidth}{0pt}{0pt}
        \vspace{-25pt}
        \centering
      \begin{nonfloat}
\centering
\vspace{-0.2cm}
\begin{minipage}{0.55\textwidth}
  \centering

  \includegraphics[width=\textwidth]{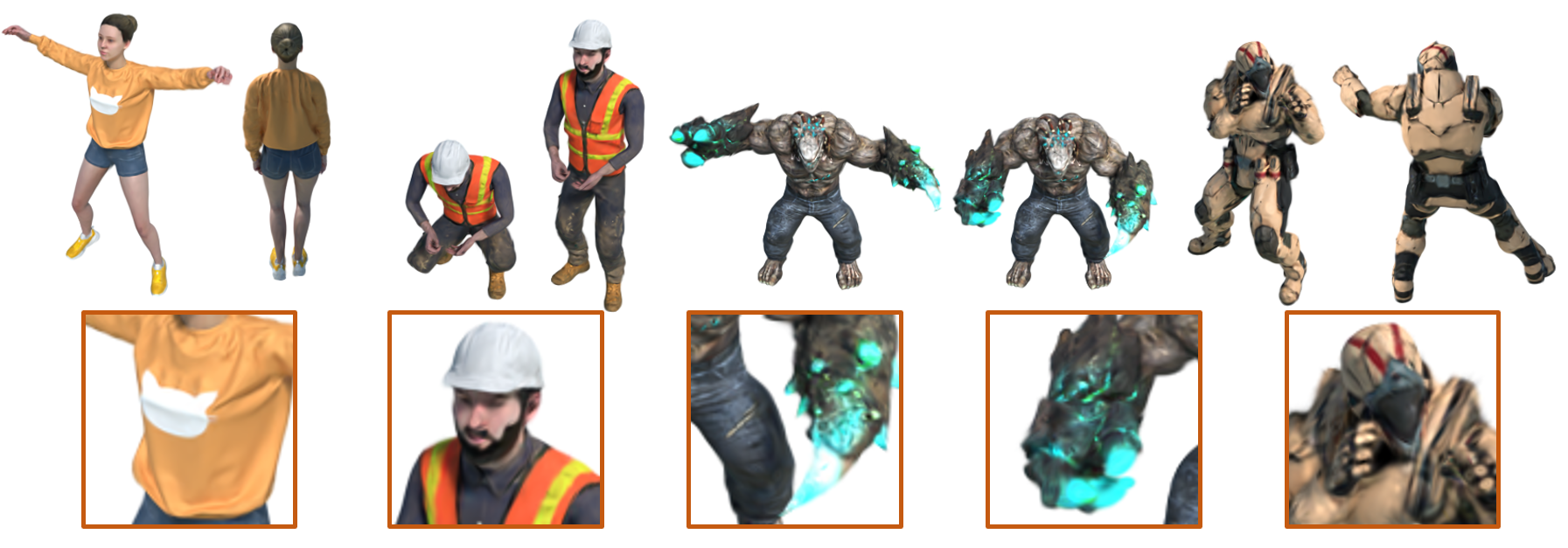}

  \textbf{a)} Novel View Synthesis on the D-NeRF Dataset
\end{minipage}\hfill
\begin{minipage}{0.421\textwidth}
  \centering
  \includegraphics[width=\textwidth]{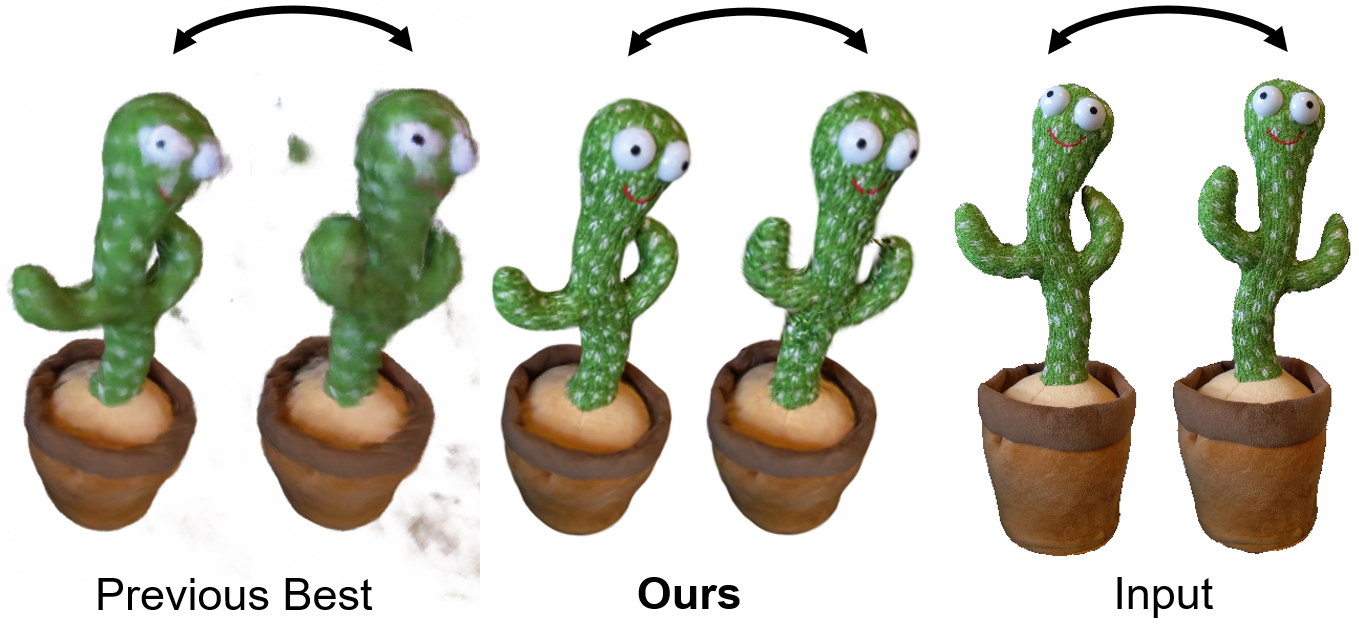}
  \textbf{b)} Comparison on an Unbiased4D Sequence.
\end{minipage}
\vspace{-0.2cm}
\captionof{figure}{We present \textbf{Neural Parametric Gaussians (NPGs)}, a method for monocular non-rigid reconstruction of objects. \textbf{a)} Our method enables to produce high quality reconstructions in easy settings like the object-level D-NeRF scenes while \textbf{b)} also being able to handle challenging monocular settings much better than previous work through strong parametric low-rank regularization.\vspace{-0.2cm}}
\label{fig:teaser}
\end{nonfloat}
    \end{adjustwidth}
\vspace{0.05cm}
\end{center}
 }
\makeatother
\maketitle

\input{sec/0_abstract}    
\input{sec/1_intro}
\input{sec/2_related_work}
\input{sec/3_method}

\input{sec/4_experiments}

\input{sec/5_limitations}
\input{sec/6_conclusion}
{
    \small
    \bibliographystyle{ieeenat_fullname}
    \bibliography{main}
}

\input{supplementary/main}

\end{document}

%% file: preamble.tex
\usepackage[dvipsnames]{xcolor}


%% file: sec/0_abstract.tex
\begin{abstract}
\vspace{-0.2cm}
\textit{Reconstructing dynamic objects from monocular videos is a severely underconstrained and challenging problem, and recent work has approached it in various directions. However, owing to the ill-posed nature of this problem, there has been no solution that can provide consistent, high-quality novel views from camera positions that are significantly different from the training views. In this work, we introduce Neural Parametric Gaussians (NPGs) to take on this challenge by imposing a two-stage approach: first, we fit a low-rank neural deformation model, which then is used as regularization for non-rigid reconstruction in the second stage. The first stage learns the object’s deformations such that it preserves consistency in novel views. The second stage obtains high reconstruction quality by optimizing 3D Gaussians that are driven by the coarse model. To this end, we introduce a local 3D Gaussian representation, where temporally shared Gaussians are anchored in and deformed by local oriented volumes. The resulting combined model can be rendered as radiance fields, resulting in high-quality photo-realistic reconstructions of the non-rigidly deforming objects. We demonstrate that NPGs achieve superior results compared to previous works, especially in challenging scenarios with few multi-view cues.\footnote[1]{Project Page: \href{https://geometric-rl.mpi-inf.mpg.de/npg/}{https://geometric-rl.mpi-inf.mpg.de/npg/}}} 
\end{abstract}

%% file: sec/1_intro.tex
\section{Introduction}
\label{sec:intro}
Reconstructing 3D objects from 2D observations is a core problem in computer vision with numerous applications in several industries, such as the movie and game industry, AR/VR, and robotics. Tremendous progress has been seen in static scene reconstruction during the last few years. The real world is, however, dynamic, and most recorded scenes are captured in a casual setting, with sparse coverage from a single camera. Thus, addressing these two aspects during reconstruction is of fundamental importance.

The success of neural approaches on static scenes has encouraged their use for dynamic scene reconstruction from monocular videos, both in its classical~\cite{cvpr_LiNSW21, cvpr_PumarolaCPM21, iccv_ParkSBBGSM21} and hybrid~\cite{cao2023hexplane, fridovich2023k, liu2022devrf} forms. These methods either learn a per-frame scene representation with limited time consistency~\cite{cvpr_LiNSW21, fridovich2023k} or utilize a time-invariant canonical space, which is used to track the observations at each timestep~\cite{cvpr_PumarolaCPM21, park2021hypernerf}. However, as pointed out in Gao \textit{et al.}~\cite{gao2022dynamic}, they are evaluated on datasets that contain multi-view signals, such as camera teleportation\textemdash i.e., alternating samples from different cameras to construct a temporal sequence\textemdash and limited object motion, and their performance suffers drastically when evaluated on more realistic monocular sequences. Such sequences usually contain faster object motion compared to camera motion. Strong regularization is required in order to propagate information between different timesteps with the correct data association. 
As we will demonstrate on realistic sequences, current methods fail to provide such regularization. Traditionally, a possible solution to this has been to use geometry proxies in different forms~\cite{johnson2022unbiased}, such as the SMPL~\cite{SMPL:2015} body model. However, these approaches require an appropriate model with a priori knowledge.

In this paper, we tackle the problem of template-free dynamic object reconstruction from realistic monocular captures. 
We propose \emph{Neural Parametric Gaussians (NPGs)} as a two-step approach, which first learns the coarse, deforming geometry and then uses it as a proxy for reconstruction. In the first step, we obtain a coarse neural parametric model for the observed object at each timestep, generated using a low-rank decomposition of deformation and a set of point basis vectors. The basis vectors force the model to share information between timesteps, thus providing regularization for the sparsely observed dynamic regions. In the second step, we represent fine-level geometry and appearance details by initializing and optimizing 3D Gaussians on top of the coarse point template, inspired by the recent breakthrough provided by the 3D Gaussian splatting approach for static scenes~\cite{kerbl20233Dgaussians}. For handling dynamic objects, we define Gaussians in oriented local volumes such that they can be driven by the coarse deformation model from the first stage.
In short, our contributions are:
\begin{itemize}
\item A two-stage reconstruction approach that learns coarse deformation in stage one and uses it as a constraint for reconstruction in stage two.
\item A coarse parametric point model based on a low-rank deformation that provides regularization and correspondences over time for object reconstruction.
\item A 3D Gaussian splatting approach, where Gaussians live in deforming local volumes, generalizing 3D Gaussians from static to dynamic scenes efficiently.
\item We show that our model improves on the state of the art of non-rigid novel view synthesis from a monocular camera, especially in challenging cases with few multi-view cues.
\end{itemize}

%% file: sec/2_related_work.tex
\section{Related Work}
\label{sec:related_work}
We look at a brief overview of the recent advances in non-rigid reconstruction from the perspective of scene representations, deformation modeling, and supervision, focusing on monocular and object-level reconstruction methods. For a more in-depth discussion, please refer to the recent survey by Yunus \textit{et al.}~\cite{yunus2024recent}.

\subsection{Neural Rendering of Dynamic Scenes}
NeRF~\cite{mildenhall2020nerf} has been extended to model dynamic scenes both using a global MLP representation and as a hybrid representation, where neural features are stored at the nodes of a discrete data structure.
These representations capture the scene dynamics in one of two ways. One approach, dubbed \textit{Space-Time Neural Fields}, directly adds an extra time dimension to the scene representation to reconstruct the dynamic scene from multi-view~\cite{li2022neural, song2022pref, attal2023hyperreel, park2023temporal, wang2022mixed, wang2022fourier, li2022streaming, wang2023neural, wang2022neus2, lin2023im4d, xu20234k4d} or monocular~\cite{cvpr_LiNSW21, cvpr_XianHK021, iccv_DuZYT021, GaoICCVDynNeRF, li2023dynibar, cao2023hexplane, fridovich2023k, shao2022tensor4d, song2023nerfplayer} video. Hybrid methods further achieve acceleration by parameterizing the 4D scene representation using voxel grids~\cite{park2023temporal, wang2022mixed, wang2022fourier, li2022streaming, wang2023neural, wang2022neus2, song2023nerfplayer} or planar factorization~\cite{attal2023hyperreel, cao2023hexplane, fridovich2023k, shao2022tensor4d, lin2023im4d, xu20234k4d}.
Such methods often rely on estimated depth maps and pre-computed optical flow for local motion modeling but fail to propagate information globally, required to render novel views from viewpoints significantly different than the observations, making them suitable mostly for forward-facing videos. Similar to ours, Hexplane~\cite{cao2023hexplane} shares information between timesteps by proposing a low-rank temporal basis for the spatially decomposed volumes. However, they optimize the basis together with the fine-level geometry and appearance while our two-stage approach provides stronger regularization by first obtaining a coarse template and then optimizing fine details on top.

The other paradigm, called \textit{Deformable NeRFs}, introduces a 4D deformation field that maps the observations at each timestep to a canonical space, i.e. backward warping, ensuring temporal correspondences in contrast to the previous paradigm but restricting large deviations and topological changes from the canonical field~\cite{cvpr_PumarolaCPM21, iccv_ParkSBBGSM21, iccv_TretschkTGZLT21, park2021hypernerf, wang2023flow, Choe_2023_ICCV}. A few approaches represent the 4D deformation and the canonical radiance field with a voxel-based hybrid representation for fast multi-view reconstruction~\cite{liu2022devrf, tretschk2023scenerflow}. TiNeuVox~\cite{TiNeuVox} further accelerates monocular reconstruction by using a very light MLP for the backward deformation field, compensating for compression by enhancing the scene representation through temporal embeddings. Backward warping is not smooth; hence, it has difficulties generalizing to sparse capture settings for novel view synthesis, where strong regularization is required. Recently, ForwardFlowDNeRF~\cite{Guo_2023_ICCV} proposed the use of time-dependent voxel features in the canonical space for learning a forward deformation field to each timestep instead of the other way around, leading the MLP to model motion more smoothly. In contrast, our optimized coarse model provides explicit regularization for reconstruction in the second stage.

\subsection{Object-level Non-Rigid Reconstruction}
Object-level approaches employ masks to separate the dynamic object of interest from the background and mostly focus on surface modeling. Using classical representations, Shape-from-Template methods~\cite{kairanda2022f, casillas2021isowarp} rely on pre-acquired full object templates, while template-free online methods~\cite{chang2023mono, lin2022occlusionfusion, bozic2020neural, newcombe2015dynamicfusion} use RGB-D input to reconstruct the object's geometry. A few recent neural rendering approaches focus on surface modeling of masked objects from monocular RGB~\cite{johnson2022unbiased} and RGB-D~\cite{cai2022neural} videos by replacing the density-based representation with SDF modeling from NeuS~\cite{wang2021neus}. Unbiased4D~\cite{johnson2022unbiased} also utilizes a mesh proxy to model large deformations.
A couple of methods~\cite{sinha2022common, tan2023distilling} employ data-driven category-level priors for shape and appearance, along with articulation~\cite{tan2023distilling} or scene flow~\cite{sinha2022common} prediction, whereas our template prediction does not rely on any learned data priors but is purely optimization-based.

Neural Parametric Models utilize an auto-decoded MLP to learn the articulation space\textemdash either represented by scene flow~\cite{palafox2021npms, palafox2022spams, mohamed2022gnpm} or volumetric bone transformations as in BANMO~\cite{yang2022banmo}\textemdash for a given shape from monocular videos. Similar to ours, KeypointTransporter~\cite{novotny2022keytr} reconstructs coarse 3D point clouds for objects, parameterized by a low-rank deformation basis. 
In contrast to such methods, which focus on coarse geometry only, our approach focuses on high-quality view synthesis using a deformable parametric model. 
A few approaches utilize either skeletons~\cite{li2022tava, noguchi2022unsupervised, liu2023hosnerf, yu2023monohuman} or surface templates~\cite{peng2021neural, xu2021h, Zhao_2022_CVPR, jiang2022neuman} from parametric models like SMPL~\cite{SMPL:2015} to guide the motion field and learn the canonical radiance field on top of such templates. While the template provides motion regularization, these approaches require a priori information about what is to be reconstructed, which does not generalize to sequences in the wild. Our coarse point template, on the other hand, is derived solely from the observations.

\begin{figure*}[ht]
\centering
\includegraphics[width=1\linewidth]{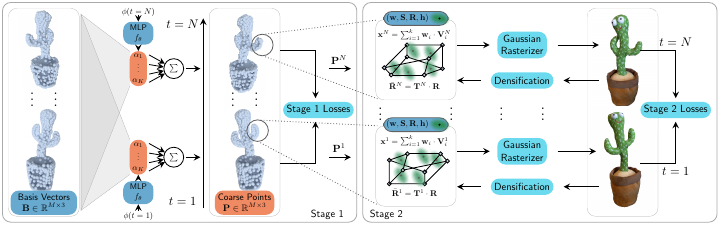}
\caption{\textbf{Overview of our method.} We present a two-stage method. In \textbf{stage 1 (left)} we learn a coarse point model, which is parameterized through low-rank coefficients from an MLP. In \textbf{stage 2 (right)}, we optimize 3D Gaussians in local volumes, defined by the point sets. The figure distinguishes between parts that are shared over time (\textcolor{shared}{$\blacksquare$}), individual for each time step (\textcolor{individual}{$\blacksquare$}), and fixed-function (\textcolor{fixed}{$\blacksquare$}). MLP weights $\theta$, Gaussian interpolation weights $\mathbf{w}$, scales $\mathbf{S}$, rotations $\mathbf{R}$ and harmonic coefficients $\mathbf{h}$ are shared over time and the deformation is purely modeled by the low-rank coefficients $\alpha_i$, leading to a different coarse point model for each frame.\vspace{-0.3cm}}
\label{fig:overview}
\end{figure*}

\subsection{Point-based Representation and Rendering}
A few approaches utilize point-based neural rendering, introduced for static scenes by Point-NeRF~\cite{Xu2022PointNeRF}, for dynamic avatar modeling from multi-view~\cite{uzolas2023template} and monocular~\cite{su2023npc} videos.
The recent trend has been shifting towards using purely point-based explicit representations for their efficiency, showing great results from multi-view input for dynamic surface modeling~\cite{prokudin2023dynamic} and view synthesis using point-based rasterization~\cite{zhang2022differentiable}. 3D-GS~\cite{kerbl20233Dgaussians}, a recent breakthrough approach for static scenes, models the scene with numerous volumetric 3D Gaussians and proposes a Gaussian splatting-based rasterization technique to deliver state-of-the-art results in terms of view synthesis quality and rendering speed. Our approach extends the static 3D-GS method to dynamic objects. 

\vspace{-0.3cm}
\paragraph{Concurrent Works.} In terms of modeling dynamic scenes with 3D Gaussians, several concurrent works have appeared on arXiv.org. Dynamic3DGS~\cite{luiten2023dynamic} utilizes dense multi-view inputs to track the rotation and position of 3D Gaussians initialized in the first frame across timesteps. Two similar works, Deformable3DGS~\cite{yang2023deformable3dgaussians} and 4DGaussianSplatting~\cite{wu20234dgaussians} introduce a canonical space where 3D Gaussians are initialized while using a neural deformation field to track the position, rotation, and scale of each Gaussian across timesteps. 4DGaussianSplatting additionally represents the deformation field using a decomposed Hexplane representation. In contrast, our approach introduces the 3D Gaussian scene representation specifically to the underconstrained setting of sparse monocular captures, by initializing the Gaussians from the underlying coarse point template and using the deformation of the template as a guide for optimization of the 3D Gaussians. Also, in contrast to some works~\cite{yang2023deformable3dgaussians, wu20234dgaussians}, we do not apply MLPs to all Gaussians in a scene, which leads to better scalability.

%% file: sec/3_method.tex
\vspace{-0.1cm}
\section{Method}
\vspace{-0.1cm}
\label{sec:method}
The presented method is a two-stage approach. First, we obtain constraints for reconstruction by fitting a neural parametric point model to monocular video sequences (c.f. Sec.~\ref{sec:method_template}). Then, the second stage uses the learned constraints from stage 1 to solve the underconstrained task of monocular, non-rigid reconstruction (c.f. Sec.~\ref{sec:method_reconstruction}). Specifically, we formulate 3D Gaussians as a function of the underlying point template, optimizing the parameters with respect to local subspaces.

\subsection{Coarse Parametric Point Model}
\label{sec:method_template}
The goal of the first stage is to obtain a coarse, non-rigid point model that follows the movement of the object, given a sequence of monocular input views $\{\mathbf{I}_i\}_{i=1}^N$ with masks $\{\mathbf{M}_i\}_{i=1}^N$ and camera poses. For real videos the poses are obtained via COLMAP~\cite{schoenberger2016sfm, 
schoenberger2016mvs}. We do not make any assumptions of 3D location of the object, but we require the object to stay in view. Note that, given recent advances in segmentation, in contrast to previous work, we consider the mask to be a weak assumption and demonstrate that our method works with masks obtained from Segment Anything (SAM)~\cite{kirillov2023segment} in the supplemental materials.
\vspace{-0.2cm}
\subsubsection{Representation}
 We represent our coarse, non-rigid model as a set of points $\mathbf{P}^t \in \mathbb{R}^{M\times3}$ for $t\in\{1,...,N\}$, which are the output of a low-rank deformation basis 
\begin{equation}
\mathbf{P}^t = \sum_{k=1}^{K} \alpha_k^t \mathbf{B}_k \text{,}
\end{equation}
where $\{\mathbf{B}_k \in \mathbb{R}^{M\times 3}\}_{k=1}^K$ is a learnable point basis and $\alpha_k^t \in \mathbb{R}$ are low-rank coefficients. The latter is produced by an MLP $f_\theta:\mathbb{R}^d \rightarrow \mathbb{R}^{K}$, mapping the temporally encoded time $\phi(t) \in \mathbb{R}^d$ to coefficients over time, following the ideas of multiple recent works ~\cite{novotny2022keytr, wewer2023simnp, palafox2021npm}. The basis is randomly initialized. This representation naturally provides correspondences over time, as the point cloud is smoothly deformed between frames. 
By choosing $K$, we have control over the rigidity of the model. Intuitively, each basis vector can represent the object in a certain pose, and the linear combination of these shapes defines the extent of deformations that can be modeled, thus providing regularization. We choose $K << T$ and evaluate different choices in Sec.~\ref{sec:ablation}. Additionally, the points are equipped with coarse colors $\mathbf{C} \in \mathbb{R}^{N\times 3}$ and random features $\mathbf{E} \in \mathbb{R}^{N\times d}$ for identification purposes. 

\vspace{-0.1cm}
\subsubsection{Optimization}
 Given a sequence of input views $\{\mathbf{I}_i\}_{i=1}^N$ and masks $\{\mathbf{M}_i\}_{i=1}^N$, we follow a similar procedure as in KeyTr~\cite{novotny2022keytr} and find optimal MLP weights $\hat{\theta}$, a deformation basis $\{\hat{\mathbf{B}}_k\}_{k=1}^K$ and colors $\mathbf{\hat{C}}$ as
\begin{equation}
   \hat{\theta}, \{\hat{\mathbf{B}}_k\},  \hat{\mathbf{C}}^t = \argmin_{\theta, \{\mathbf{B}_k\}, \mathbf{C}^t} \,\, (\mathcal{L}_\text{M} + \mathcal{L}_\text{R}+  \mathcal{L}_\text{OF}) \text{,}
\end{equation}
with the individual losses being described in the following. For the loss weights, please refer to the supplementals.

\vspace{-0.3cm}
\paragraph{Mask Loss.} 
For each frame $i$, we sample points $\mathbf{P}_\text{M}^i$ uniformly from the given mask and minimize the Chamfer distance $\mathcal{L}_\text{M} = \text{CD}(\mathbf{P}_\text{M}^i, \pi(\mathbf{P}^i))$ to the point model projected to the image plane by $\pi$. We found this to be more efficient and equally good as a Sinkhorn matching loss~\cite{novotny2022keytr}.

\vspace{-0.3cm}
\paragraph{Rigidity Loss.} 
We add a rigidity loss that keeps neighborhood point distances of a frame $t$ similar to that of a reference frame $r$:
\begin{equation}
\mathcal{L}_\text{R} = \sum_{i=2}^M \sum_{j \in \mathcal{N}(i)} \norm{\norm{\mathbf{P}_i^r - \mathbf{P}_j^r}_2 - \norm{\mathbf{P}_i^{t} - \mathbf{P}_j^{t}}_2}_2^2 \text{,}
\end{equation}
for all $t \in \{1, ..., N\}$, where $\mathcal{N}(i)$ denotes the neighborhood points of point $i$, given a radius criterion, and $r$ is a reference frame selected manually from the dataset.


\vspace{-0.3cm}
\paragraph{Optical Flow Loss.} We use RAFT~\cite{teed2020raft} as an off-the-shelf optical flow estimator to align point movement between frames via estimated optical flow, as done by KeyTr~\cite{novotny2022keytr}. The flow consistency loss between two frames is given as
\begin{equation}\label{eq:Temporal consistency}
\begin{aligned}
    \mathcal{L}^{t\mapsto t+1}_{\text{OF}} = \norm{ \mathbf{M}^t \odot [R(\mathbf{P}^t,\mathbf{E})-R^{B}(\mathbf{P}^{t+1},\mathbf{E})])}_{\epsilon}
\end{aligned}
\end{equation}
\text{where} $R(\mathbf{P}^t,\mathbf{E})$ renders the point cloud $\mathbf{P}^t$ with random descriptors $\mathbf{E}$, and $R^{B}(\mathbf{P}^{t+1},\mathbf{E})$ renders, applies the backward optical flow and bilinear samples the result. $\norm{x}_\epsilon$ is an element-wise Huber loss with threshold $\epsilon=0.01$.

\subsection{Neural Parametric Gaussians}
\label{sec:method_reconstruction}
Given our parametric point model, we perform a detailed reconstruction of the monocular sequence by letting the coarse model drive 3D Gaussians that live in oriented local volumes defined by point sets. We first define our local volumes with Gaussians in Sec.~\ref{sec:local_volumes}. Then, the rendering process is described in Sec.~\ref{sec:rendering}, the densification process in Sec.~\ref{sec:densification}, and the optimization process in Sec.~\ref{sec:optimization}.

\subsubsection{Representation}
\label{sec:local_volumes}
When defining the local volumes, the goal is to make as few assumptions about the object as possible and, particularly, to avoid the assumption of a surface. Thus, we consider local unstructured point sets as volumes.

\vspace{-0.3cm}
\paragraph{Local Volumes.} We define an \emph{oriented local volume} as $ (\mathbf{V}(i), \mathbf{T}(i))$, where $\mathbf{V}(i)~\in \mathbb{R}^{k\times 3}$ is the point with index $i$ plus its $k-1$ nearest neighbors and $\mathbf{T}(i)\in SO(3)$ is the local volume orientation. We choose $k=20$ in our experiments and the local orientation $\mathbf{T}(i)$ heuristically by taking the orientation of the triangle of point $i$ and its closest 2 neighbors. The neighbors defining the triangle are fixed over time (please refer to the supplemental materials for a detailed description). 
Then, the full set of local volumes at time $t$ is given as
\begin{equation}
    \mathcal{V} = \{(\mathbf{V}^t(i), \mathbf{T}^t(i)) \mid \forall i \in \{1, ..., M\} \} \text{.}
\end{equation}
The set of volumes is fixed over time, however, their point positions $\mathbf{V}^t(i)$ change from frame to frame. Note that the absolute orientation of the local volumes is not relevant, as the rotations $\mathbf{R}$ of individual Gaussians can adjust accordingly (see next paragraph). Importantly, the $\mathbf{T}(i)$ are equivariant to local deformation: if the local point set is rotated, the rotation-defining triangle rotates, resulting in $\mathbf{T}(i)$ changing by the same rotation, as desired. 

\vspace{-0.3cm}
\paragraph{3D Gaussians.} Each local volume can store an arbitrary number of 3D Gaussians~\cite{kerbl20233Dgaussians}, which we denote as $\mathcal{G}_i$. Having a different number of Gaussians for each volume allows the model to adapt to the amount of high-frequency details in individual object parts.  A single Gaussian $g\in \mathcal{G}_i$ is defined as tuple $g = (\mathbf{w}, \mathbf{S}, \mathbf{R}, \mathbf{h})$, where $\mathbf{S}$, $\mathbf{R}$, and $\mathbf{h}$ are scale matrix, rotation matrix and spherical harmonic coefficients as in 3D Gaussian splatting~\cite{kerbl20233Dgaussians}, respectively. Instead of representing the 3D Gaussian position in world space, we represent its position as a set of barycentric interpolation weights $\mathbf{w}\in \mathbb{R}^k$ applied to the neighboring points $\mathbf{V}(i)$. Before rendering a Gaussian belonging to volume $i$, we find the final Gaussian position $\mathbf{x}^t$ and rotations $\bar{\mathbf{R}}^t $ at time $t$:
\begin{equation}
    \mathbf{x}^t = \sum_{j=1}^k \text{Softmax}(\mathbf{w})_j\cdot \mathbf{V}^t(i)_j \text{,}\,\,\, \text{and} \,\,\, \bar{\mathbf{R}}^t = \mathbf{T}^t(i) \cdot\mathbf{R} \text{,}
\end{equation}
via a linear combination of volume-defining points and rotation into the local coordinate frame, respectively. The process can be understood as a simplified variant of cage-based deformation transfer~\cite{ju2005cage}. It is important to note that Gaussian parameters are shared across all $t$ and that temporal changes are modeled purely by the coarse point set.

\input{tables/synthetic_dataset}

\vspace{-0.2cm}
\subsubsection{Rendering}
\label{sec:rendering}
The sets of Gaussians from all volumes in $\mathcal{V}$ are rendered using the differentiable rasterizer provided by Kerbl \textit{et al.}~\cite{kerbl20233Dgaussians}.
It is important to note that, in contrast to concurrent works~\cite{yang2023deformable3dgaussians, wu20234dgaussians}, our model does not apply networks to Gaussians. Since the number of Gaussians within one scene can become very large, this avoids heavy runtime increases for larger scenes. We only have a network driving our initial point template, which has a comparably low number of points, while Gaussians just follow their local movement and have directly optimizable parameters that are shared over the full sequence. Further, given an optimized representation, it is straightforward to extract all Gaussians and store them for real-time rendering.

\vspace{-0.2cm}
\subsubsection{Gaussian Initialization and Densification}
\label{sec:densification}
When initializing our model, we fill each volume with a small number of Gaussians. During training, we apply densification and pruning procedures from Kerbl \textit{et al.}~\cite{kerbl20233Dgaussians}. Importantly, when splitting a Gaussian, we assign new Gaussians to the same volume as the original Gaussian and apply a tiny bit of noise to weights $\mathbf{w}$ of both. In practice, we store all Gaussians in a global list and keep a volume index for each Gaussian, allowing different amounts of Gaussians for each volume. 

\vspace{-0.2cm}
\subsubsection{Optimization}
\label{sec:optimization}
In stage 2, we mainly optimize the individual Gaussians. Thus, we find optimal weights $\hat{\mathbf{w}}$, rotations $\hat{\mathbf{R}}$, scales $\hat{\mathbf{S}}$ and spherical harmonic coefficients $\hat{\mathbf{h}}$ as
\begin{equation}
\hat{\mathbf{w}}, \hat{\mathbf{R}}, \hat{\mathbf{S}}, \hat{\mathbf{h}} = \argmin_{\mathbf{w}, \mathbf{R}, \mathbf{S}, \mathbf{h}} \quad (1- \lambda)\mathcal{L}_1 + \mathcal{L}_\text{D-SSIM} \text{,}
\end{equation}
where the loss functions are $l1$-distance and a structural similarity loss, as used by Kerbl \textit{et al.}~\cite{kerbl20233Dgaussians}. Optionally, we can finetune all parameters from stage 1 based on these rendering losses, including a weak regularization to prevent large changes. While this is not strictly necessary, it slightly improves the details of the resulting reconstruction.

%% file: tables/synthetic_dataset.tex
\begin{table*}[t]
\small
\centering

\scalebox{0.92}{
\begin{tabular}{l|ccc|ccc|ccc|ccc}
\toprule
& \multicolumn{3}{c|}{Jumping Jacks} & \multicolumn{3}{c|}{Hell Warrior} & \multicolumn{3}{c|}{Hook} & \multicolumn{3}{c}{Stand Up} \\ 
Method & PSNR$\uparrow$ & SSIM$\uparrow$ & LPIPS$\downarrow$ & PSNR$\uparrow$ & SSIM$\uparrow$ & LPIPS$\downarrow$ & PSNR$\uparrow$ & SSIM$\uparrow$ & LPIPS$\downarrow$& PSNR$\uparrow$ & SSIM$\uparrow$ & LPIPS$\downarrow$ \\ 
\midrule
D-NeRF~\cite{cvpr_PumarolaCPM21} & 32.80 & 0.98 & \textbf{0.03} & 25.02 & 0.95 & 0.06 & 29.25 & 0.96 & 0.11  & 32.79 & 0.98 & \textbf{0.02}\\ 
TiNeuVox ~\cite{TiNeuVox} & \textbf{34.23} & 0.98 & \textbf{0.03} & 28.17 & 0.97 & 0.07 & 31.45 & 0.97 & 0.05 & 35.43 & \textbf{0.99} & \textbf{0.02} \\ 
HexPlane~\cite{cao2023hexplane} & 31.65 & 0.97 & 0.04 & 24.24 & 0.94 & 0.07 & 28.71 & 0.96 & 0.05 & 34.36 & 0.98 & \textbf{0.02}\\
\rowcolor{rowhighlight}%
\textbf{Ours} & 34.06 & \textbf{0.99} & \textbf{0.03} & \textbf{38.73} & \textbf{0.98} & \textbf{0.04} & \textbf{33.73} & \textbf{0.98} & \textbf{0.03} & \textbf{37.95} & \textbf{0.99} & \textbf{0.02}\\

\toprule
& \multicolumn{3}{c|}{Mutant} & \multicolumn{3}{c|}{T-Rex} & \multicolumn{3}{c|}{Lego} & \multicolumn{3}{c}{Average} \\ 
Method & PSNR$\uparrow$ & SSIM$\uparrow$ & LPIPS$\downarrow$ & PSNR$\uparrow$ & SSIM$\uparrow$ & LPIPS$\downarrow$ & PSNR$\uparrow$ & SSIM$\uparrow$ & LPIPS$\downarrow$ & PSNR$\uparrow$ & SSIM$\uparrow$ & LPIPS$\downarrow$ \\ 
\midrule
D-NeRF~\cite{cvpr_PumarolaCPM21} & 31.29 & 0.97 & \textbf{0.02} & 31.75 & 0.97 & 0.03 & 21.64 & 0.83 & 0.16 & 29.22 & 0.95 & 0.06 \\ 
TiNeuVox ~\cite{TiNeuVox} & 33.61 & 0.98 & 0.03 & \textbf{32.70} & \textbf{0.98}& 0.03 & 25.02 & 0.92 & 0.07 & 31.52 & 0.97 & 0.04\\ 
HexPlane~\cite{cao2023hexplane} & 33.79 & 0.98 & 0.03 & 30.67 & \textbf{0.98} & 0.03 & \textbf{25.22} & \textbf{0.94} & \textbf{0.04} & 29.81 & 0.96 & 0.04\\ 
\rowcolor{rowhighlight}%
\textbf{Ours} & \textbf{35.82} & \textbf{0.99} & \textbf{0.02} & 32.35 & \textbf{0.98} &  \textbf{0.02} & 24.82 & 0.93 & 0.05 & \textbf{33.92} & \textbf{0.98} & \textbf{0.03} \\
\bottomrule
\end{tabular}}\vspace{-0.2cm}
\caption{\textbf{Novel view synthesis on the synthetic D-NeRF dataset.} We evaluate our method quantitatively on scenes from the synthetic D-NeRF dataset, which fit our object-centric setting. It can be seen that our NPGs reach state-of-the-art performance. Even if the differences in metrics are small, the qualitatively results in Fig.~\ref{fig:qualitative_dnerf} show clear differences in level of detail and correspondences. The Lego and T-Rex sequences pose a special challenge to our object-level method, as they contain a large static ground plane. We obtain results on these sequences by obtaining masks for dynamic objects only via Segment-Anything~\cite{kirillov2023segment}, as detailed in the supplementary materials.}
\label{tab:synthetic_dataset}
\vspace{-0.3cm}
\end{table*}

%% file: sec/4_experiments.tex
\begin{figure}[t]
\centering
\includegraphics[width=0.5\textwidth]{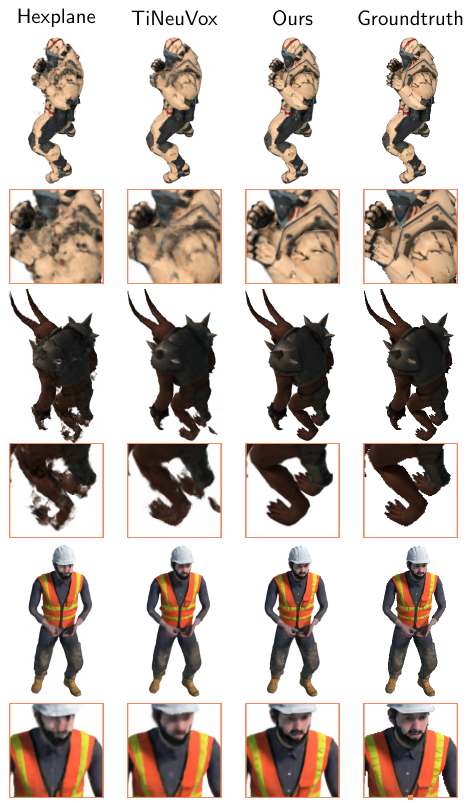}
\caption{\textbf{Qualitative comparison on novel views of the D-NeRF dataset.} We can see that our method produces more detailed reconstruction than previous work. Also, even with multi-view cues in D-NeRF, previous methods fail to always keep correct correspondences, as seen in the second example around the feet. In contrast, our NPGs keeps the shape coherent at all times and captures high frequency details under deformation. \vspace{-0.2cm}}
\label{fig:qualitative_dnerf}
\end{figure}
\section{Experiments}
\label{sec:experiments}
In this section, we evaluate our method on synthetic and real monocular datasets and perform ablation studies to demonstrate its effectiveness in reconstructing realistically captured dynamic scenes.

\vspace{-0.3cm}
\paragraph{Datasets.}
First, we evaluate our method on the commonly used D-NeRF benchmark dataset~\cite{cvpr_PumarolaCPM21} and show that it reaches state-of-the-art performance on objects which fit our setting. Note that the dataset contains unrealistic multi-view cues, like camera teleportation, which aid the reconstruction. After that, we provide quantitative, qualitative, and ablation studies on the more realistically captured Unbiased4D dataset~\cite{johnson2022unbiased} to demonstrate that our method provides adequate regularization for this underconstrained setting, where a camera is moving slowly around an object. We use the Effective Multi-view Factor (EMF) metric from Gao \textit{et al.}~\cite{gao2022dynamic} to determine the amount of multi-view cues present in datasets. D-NeRF has an EMF $\omega$ value of $2135.45$, while the average EMF $\omega$ of Unbiased4D is $24.93$. 

\vspace{-0.3cm}
\paragraph{Comparison Methods.}
Existing object-level methods mostly focus on geometry reconstruction and neglect appearance. So, we choose state-of-the-art methods for monocular scene-level reconstruction and use object masks for comparison. We mainly compare against Hexplane~\cite{cao2023hexplane}, which uses planar decomposition and information sharing for the 4D volume, and TiNeuVox-B~\cite{TiNeuVox}, which utilizes a voxel-parametrized hybrid field for deformation modeling. We use the base version (B) of TiNeuVox, which is a bit slower but more accurate than the small version.
Further comparisons are made against HyperNeRF~\cite{park2021hypernerf}, Nerfies~\cite{iccv_ParkSBBGSM21} and D-NeRF~\cite{cvpr_PumarolaCPM21} to put our performance more in context. We report the PSNR, SSIM~\cite{wang2004image}, and LPIPS~\cite{zhang2018perceptual} metrics for all experiments to quantitatively evaluate the view synthesis quality.

\vspace{-0.3cm}
\paragraph{Implementation Details.}
For stage one optimization, we use a learning rate of $0.0005$ with a cosine annealing scheduler and warm-up, a batch size of 10, the Adam optimizer~\cite{kingma2014adam} and optimize for a total of 100k iterations. We employ 1500 points to model the coarse object geometry and a 6-layer MLP to obtain the coefficients for the deformation basis.
In stage two, we mainly adopt the learning rate schedule from 3DGS~\cite{kerbl20233Dgaussians}. The LR for interpolation weights $\mathbf{w}$ is the same as the default learning rate for point position as used in 3DGS, while we adapt the learning rate of the Gaussian scale to 0.0005 to match the smaller scenes. We train for a total of 70k iterations and use an exponential learning rate scheduler for the Gaussian scales, with a very high rate initially and a gradual decay to $0.0005$ in the end. For finetuning stage 1, we use a learning rate of 0.0001.

\subsection{Results and Comparison}
\begin{figure}[t]
\centering
\includegraphics[width=0.47\textwidth]{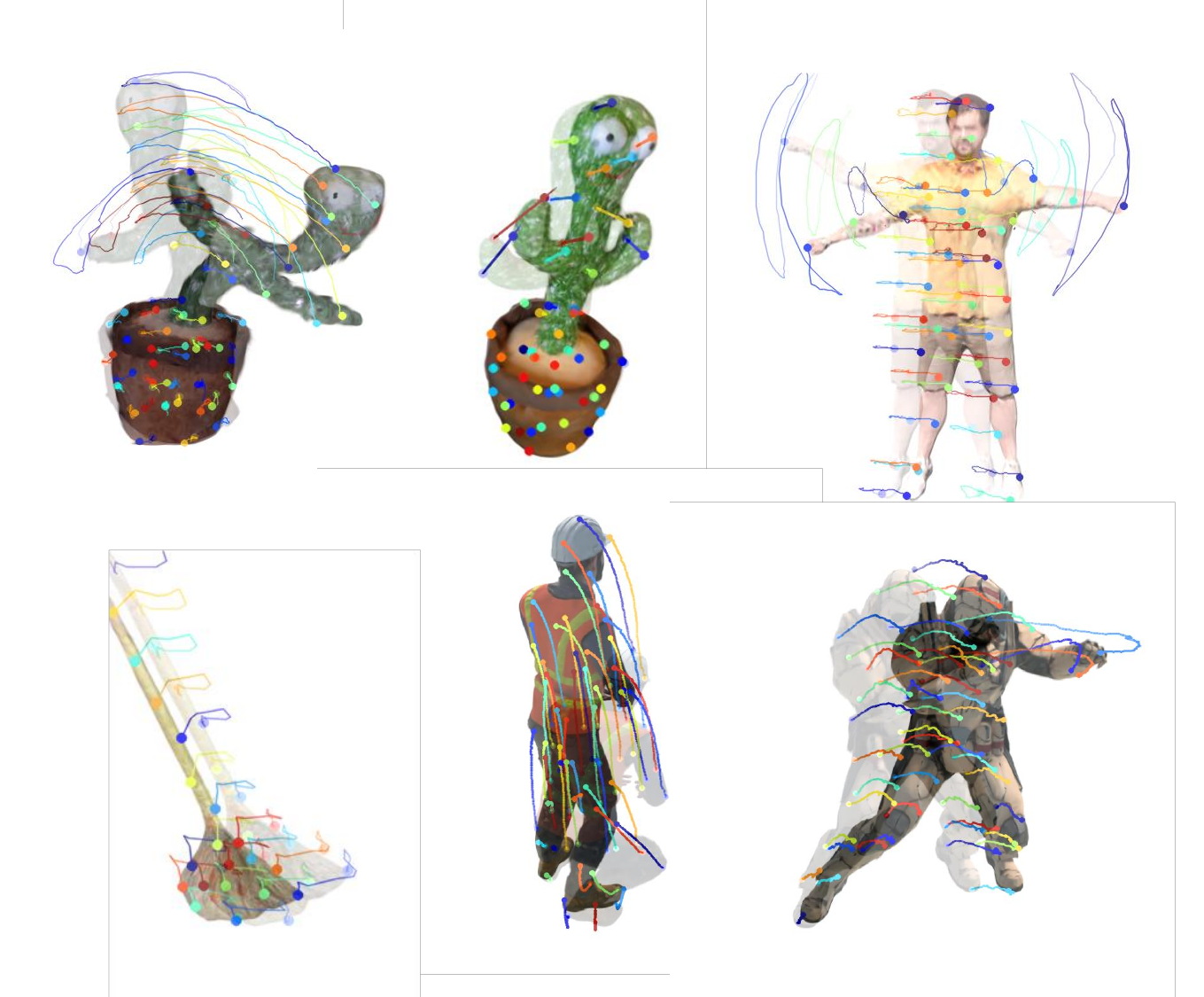}
\vspace{-0.7cm}
\caption{\textbf{Point trajectory visualization.} Our coarse parametric model automatically provides point trajectories, which in turn demonstrates the quality and smoothness of our optimized templates here. \textbf{Top Row:} Synthetic Cactus, Real Cactus and Synthetic Human sequences from the Unbiased4D dataset. \textbf{Bottom Row:} Jumping Jack, Stand Up and Hook sequences from the D-NeRF dataset. Note that the human on the top right is sliding with constant speed in this sequence, which is visible in the trajectories.}\vspace{-0.4cm}
\label{fig:trajectory}
\end{figure}

\paragraph{Comparison on D-NeRF Dataset.}
The D-NeRF dataset~\cite{cvpr_PumarolaCPM21} is a monocular dataset with 360 views of synthetic objects. It provides eight sequences with 200 training views and 20 test views. The sequences contain teleporting camera motion, which provides multi-view cues for the reconstruction, relaxing the difficulty  for the dataset~\cite{gao2022dynamic}. We compare renderings with resolution $400 \times 400$ as done by previous work and provide results on the full resolution of $800 \times 800$ in the supplemental materials. Although D-NeRF is not our target setting because of strong multi-view cues, it provides verification that our method can reconstruct highly detailed objects from high-quality observations. We provide quantitative results for scenes that fit our object-level setting in Tab.~\ref{tab:synthetic_dataset}.  The results show that our method slightly outperforms the previous state-of-the-art approaches. A qualitative comparison is shown in Fig.~\ref{fig:qualitative_dnerf} and the quality of our optimized templates can be seen in the bottom row on Fig.~\ref{fig:trajectory}. Since strong regularization is not required for these sequences, most of the gains come from improved details in representation. However, as seen in the second example, there are also some cases where our method prevents inconsistencies in geometry, where previous methods fail to do so. It should also be noted that HexPlane and TiNeuVox are capable of representing high-quality details in theory and that blurriness is a result of non-rigid deformation. This suggests that our method deals better with representing high-frequency details under deformation, even with stronger regularization.
\input{tables/real_dataset}

\begin{figure*}[t]
\centering
\includegraphics[width=\textwidth]{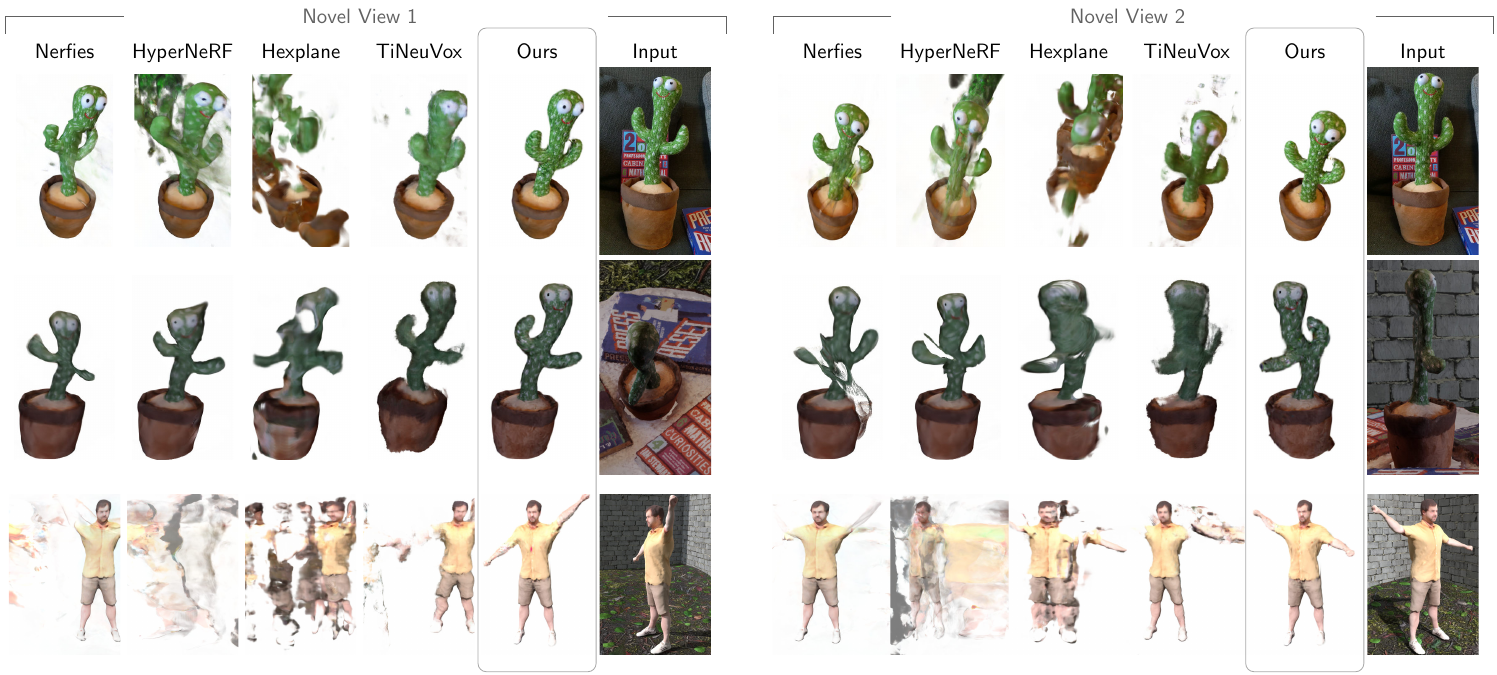}\vspace{-0.3cm}
\caption{\textbf{Qualitative comparison on the Unbiased4D dataset.} The Unbiased4D dataset is a challenging monocular video dataset where the amount of multi-view cues is very low. We show two different novel views for each scene. It can be seen that previous methods fail to correctly reconstruct the objects, while our NPGs keep the coarse geometry intact and provide novel views with high frequency details. On the most right, we show the input view for the given timeframe. Our method provides consistent reconstructions of far away novel views.\vspace{-0.5cm}}
\label{fig:qualitative_ub4d}
\end{figure*}

\vspace{-0.3cm}
\paragraph{Comparison on Unbiased4D Dataset.}
To evaluate realistically captured sequences that require strong regularization, we utilize the monocular dataset provided by Unbiased4D~\cite{johnson2022unbiased}. The original paper required geometry proxies to successfully reconstruct the sequences. It provides five sequences of both synthetic and real objects, each providing 150 training views. We use masks for us and all baselines to ensure a fair comparison. Only one of the sequences, the synthetic cactus, provides ground truth novel views for evaluating reconstruction quality quantitatively. We show the results in Tab.~\ref{tab:real_dataset}. It can be seen that we outperform all previous methods. The reasons for this can be seen in Fig.~\ref{fig:qualitative_ub4d}. While previous methods fail to capture the geometry correctly, NPGs keep it intact over the sequence and successfully fit details. The coarse point model learned in the first stage successfully serves as a proxy for reconstruction in the second stage, resolving ambiguities. The smoothness of our motion modelling can be seen in Fig.~\ref{fig:trajectory}. It is apparent that, in contrast to baselines, our model performs correct tracking, which is important for consistent reconstruction.

\begin{figure}[ht!]
\centering
\includegraphics[width=1\linewidth]{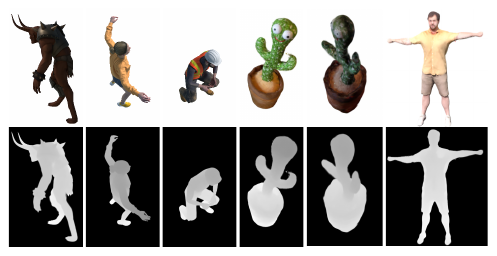}\vspace{-0.3cm}
\caption{\textbf{Rendered depth from optimized NPGs.} We render depth maps from optimized models, showing consistent geometry.\vspace{-0.1cm}}
\label{fig:ablation-basis-vectors}
\end{figure}
\input{tables/ablation}

\begin{figure}[ht]
\centering
\includegraphics[width=1\linewidth]{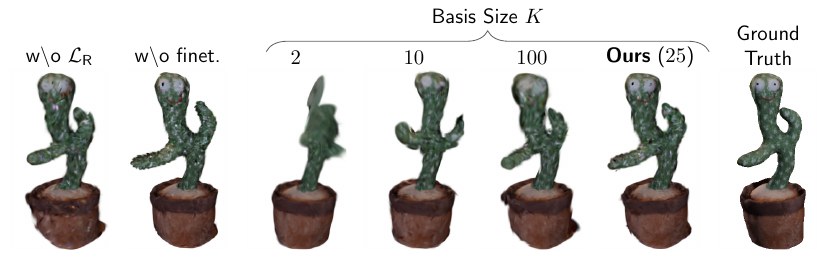}\vspace{-0.2cm}
\caption{\textbf{Qualitative ablation study on the Unbiased4D dataset.} We show ablation on a novel view of the synthetic cactus sequence at a particular timestep. The regularization provided by our low-rank deformation basis can be clearly seen with respect to the ground truth novel view on the right.}\vspace{-0.5cm}
\label{fig:ablation-basis-vectors}
\end{figure}

\vspace{-0.2cm}
\paragraph{Rendering speed} Due to the efficient Gaussian splatting rasterizer and no MLPs applied per Gaussian, NPGs have very competitive rendering times. On an NVidia A40, we render a frame in approximately $0.05s$ (20fps) in $400 \times 400$ image resolution. In comparison, TiNeuVox requires $2.15s$ per frame and Hexplane $0.94s$ per frame. Our rendering speed can be further increased by extracting the Gaussian representation explicitly and utilizing the real-time rendering of 3D Gaussian splatting~\cite{kerbl20233Dgaussians}.

\subsection{Ablation Study}
\label{sec:ablation}
\paragraph{Coarse Parametric Model Optimization.}
We provide ablation for two aspects of our coarse parametric point model optimization quantitatively in Tab.~\ref{tab:ablation} and qualitatively in Fig.~\ref{fig:ablation-basis-vectors}. First, we show the effect of utilizing the rigidity loss during stage 1 optimization, which helps keep the template consistent across timesteps. Next, we demonstrate the effect of fine-tuning stage 1 parameters during stage 2 optimization, which, while not necessary, improves the accuracy of the template nonetheless.

\vspace{-0.3cm}
\paragraph{Deformation Basis Rigidity.} We compare the effect of different regularization strengths provided by the low-rank deformation basis in Tab.~\ref{tab:ablation} and Fig.~\ref{fig:ablation-basis-vectors}, by controlling the basis size $K$. A smaller basis size over-regularizes the reconstruction, restricting non-rigid deformations, while a larger basis makes the model overfit to the observations more, hindering novel view synthesis. We find the optimal size to be 25 for the synthetic cactus sequence.

%% file: tables/real_dataset.tex
\begin{table}[t]
\small
\centering

\setlength{\tabcolsep}{3.85pt}

\begin{tabular}{l|ccc}
\toprule 
Method & PSNR$\uparrow$ & SSIM$\uparrow$ & LPIPS$\downarrow$   \\ 
\midrule
Nerfies~\cite{iccv_ParkSBBGSM21} & 17.36 & 0.87 & 0.13  \\ 
HyperNeRF~\cite{park2021hypernerf} & 18.56 & 0.88 & 0.12  \\ 
TiNeuVox ~\cite{TiNeuVox} & 15.92 & 0.848 & 0.155  \\ 
HexPlane~\cite{cao2023hexplane} & 16.327 & 0.85 & 0.16  \\ 
\rowcolor{rowhighlight}%
\textbf{Ours} & \textbf{22.348} & \textbf{0.905} & \textbf{0.095}\\
\bottomrule
\end{tabular}\vspace{-0.2cm}
\caption{\textbf{Novel view synthesis on the Unbiased4D dataset.} We clearly outperform previous methods on this challenging dataset, where the amount of multi-view cues is very low.}
\label{tab:real_dataset}
\vspace{-0.5cm}
\end{table}

%% file: tables/ablation.tex
\begin{table}[h]
\small
\centering

\begin{tabular}{l|ccc}
\toprule
Method & PSNR$\uparrow$ & SSIM$\uparrow$ & LPIPS$\downarrow$\\ 
\midrule
Ours w/o Rigidity Loss & 20.73 & 0.89 & 0.11 \\ 
Ours w/o Template Fine Tuning & 20.92 & 0.89 & 0.097 \\ 
Ours ($K=100$) & 21.80 & 0.89 & 0.10 \\
Ours ($K=10$) & 16.63 & 0.86 & 0.14 \\ 
Ours ($K=2$) & 15.67 & 0.87 & 0.15 \\ 
\rowcolor{rowhighlight}%
\textbf{Ours ($K=25$)} & \textbf{22.348} & \textbf{0.905} & \textbf{0.095} \\
\bottomrule
\end{tabular}\vspace{-0.2cm}
\caption{\textbf{Ablation study on the synthetic cactus sequence from Unbiased4D dataset.} The first two rows show the analysis for coarse parametric model optimization while the next three rows demonstrate the effect of deformation basis size $K$.}
\label{tab:ablation}
\vspace{-0.3cm}
\end{table}

%% file: sec/5_limitations.tex
\begin{figure}[t]
  \centering
  \includegraphics[width=\linewidth]{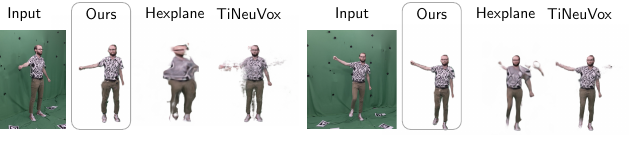}\vspace{-0.5cm}
   \caption{Failure case of a flat template and wrong deformation.\vspace{-0.6cm}}
   \label{fig:human}
\end{figure}
\paragraph{Limitations.}
\label{sec:limitations}
 We tackle the highly ill-posed problem of reconstructing \textbf{non-rigid} objects from \textbf{monocular} videos. Thus, there are limitations depending on the complexity of sequences, e.g., w.r.t. camera movement, lighting, speed and extent of deformations. Fig.~\ref{fig:human}, for example, shows results where the template collapsed to a flat surface, which we attribute to the camera being very static.

%% file: sec/6_conclusion.tex
\vspace{-0.1cm}
\section{Conclusion}
\vspace{-0.1cm}
\label{sec:conclusion}
We introduce Neural Parametric Gaussians, a high-fidelity view synthesis approach for objects captured with monocular camera. Our two-stage optimization obtains a coarse parametric point model based on a low-rank deformation basis, providing strong regularization for consistent novel-view synthesis from sparse observations, while the use of 3D Gaussians in conjunction with the template enables modeling fine geometric and appearance details efficiently. NPG is the first approach that employs the regularization power of neural parametric models for high-quality novel view synthesis. We demonstrated that, in contrast to previous work in monocular non-rigid reconstruction, we can achieve consistent models in challenging settings.

\paragraph{Acknowledgments.} R. Yunus was supported by the Max Planck \& Amazon Science Hub. Thanks to all supporters.

%% file: supplementary/main.tex
\clearpage
\setcounter{page}{1}
\setcounter{section}{0}
\renewcommand\thesection{\Alph{section}}
\maketitlesupplementary
 
In the following, we go over implementation details in Sec.~\ref{sec:other_details}, provide a discussion about using Segment Anything (SAM) masks in Sec.~\ref{sec:sam}, provide additional comparisons in Sec.~\ref{sec:more-comparisons}, provide a qualitative analysis of coarse point sets in Sec.~\ref{sec:point-templates}, and additional ablation studies in Sec.~\ref{sec:more-ablation-studies}. We also provide a \textbf{supplementary video} in addition to this document, showing results in motion.

\section{Implementation Details}
\label{sec:other_details}
In the following, we go over the network architecture, the procedure to obtain the orientations of local volumes, and additional hyperparameters, such as deformation basis size and loss weights.
\subsection{Network Architecture}
\begin{figure}[t]
\centering
\includegraphics[width=0.5\textwidth]{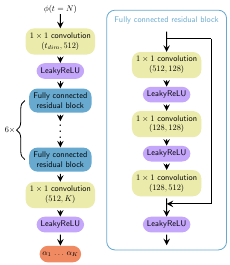}
\caption{\textbf{Basis coefficient prediction network architecture.} $t_{dim}$ is the dimension of the temporal embedding output by positional encoder $\phi$ for timestep $N$, $\alpha_i$ is the basis coefficient and $K$ is the basis size.\vspace{-0.2cm}}
\label{fig:network_architecture}
\end{figure}

Fig.~\ref{fig:network_architecture} shows the architecture for our basis coefficient prediction network. It takes the index of the current timestep, positionally encoded using Fourier features with 6 base frequencies~\cite{mildenhall2020nerf}, and outputs the coefficients for the $K$ vectors of our deformation basis for time $N$. The default scaling of $0.01$ is used for the LeakyReLU layer. 

\begin{figure}[t]
\centering
\includegraphics[width=0.5\textwidth]{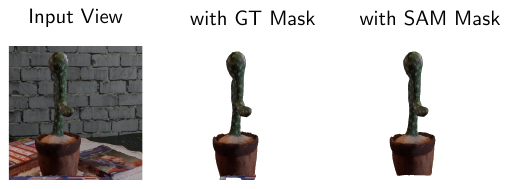}
\caption{\textbf{SAM vs. Ground truth mask.} Shown are the masked input views. It can be seen that in the case of synthetic cactus, the masks obtained using our SAM procedure are even slightly better than the GT masks. The provided masks of the dataset sometimes contain parts of occluding objects, such as the books below.}
\label{fig:sam_comparison}
\end{figure}
\input{supplementary/tables/sam_comparison}

\subsection{Orientations of Local Volumes}
We find the orientation $\mathbf{T}$ heuristically from point $i$ and its closest to neighbors $j$ and $k$. Let $\mathbf{p}_i$, $\mathbf{p}_j$ and $\mathbf{p}_k$ denote the point positions at one specific time. Then, we find the three vectors of the rotation matrix as:
\begin{align}
\mathbf{T}_{1,\cdot} &= \frac{(\mathbf{p}_2 - \mathbf{p}_1)}{\norm{\mathbf{p}_2 - \mathbf{p}_1}_2} \times \frac{\mathbf{p}_3 - \mathbf{p}_1}{\norm{\mathbf{p}_3 - \mathbf{p}_1}_2} \text{,}\\
\mathbf{T}_{2,\cdot} &= \frac{(\mathbf{p}_1 + \mathbf{p}_2 + \mathbf{p}_3)/3 - \mathbf{p}_1}{\norm{(\mathbf{p}_1 + \mathbf{p}_2 + \mathbf{p}_3)/3 - \mathbf{p}_1}_2} \text{,}\\
\mathbf{T}_{3,\cdot} &= \mathbf{T}_{1,\cdot} \times \mathbf{T}_{2,\cdot} \,\,\text{.}
\end{align}
We take the normal vector of the arising triangle as first vector, the vector pointing from triangle center to $\mathbf{p}_1$ as second vector, and find the third vector as cross product of the first two. While the absolute orientation arises from the given three points arbitrarily, one important property is ensured: if $\mathbf{p}_1$, $\mathbf{p}_2$, $\mathbf{p}_3$ are rotated by a matrix $\mathbf{R}$, i.e. $\mathbf{R}\mathbf{p}_1$, $\mathbf{R}\mathbf{p}_2$, $\mathbf{R}\mathbf{p}_3$, the resulting orientation matrix rotates by the same amount, i.e. $\mathbf{R}\mathbf{T}$ (equivariance).

\input{supplementary/tables/d-nerf_high_res}

\subsection{Deformation Basis Size}
For the D-NeRF dataset, we use a basis size of 47 for \textit{Jumping Jack}, \textit{Stand Up}, \textit{Hell Warrior}, \textit{Hook} and \textit{Mutant} sequences. For \textit{Lego} and \textit{T-Rex} sequences, we use a basis size of 30. For the Unbiased4D dataset, we use a basis size of 25 for \textit{Real Cactus} and \textit{Synthetic Cactus} sequences while the basis size for the \textit{Synthetic Human} is kept at 47.
\subsection{Loss Weightings} 
During stage one of our optimization, we use the following loss weights: for the mask loss, we use a weight of $5$ for the D-NeRF dataset and $1$ for the Unbiased4D dataset; for the optical flow loss, we use a weight of $10$ across both datasets; for the rigidity loss, we use a weight of $1e+9$ for the Unbiased4D dataset and $5e+7$ for the D-NeRF dataset. In stage two of our optimization, we introduce a weak regularization for the template fine-tuning step by keeping the points close to the original template point positions obtained from stage one. Here, we use a regularization strength of $0.1$.


\section{Segment Anything (SAM) Masks}
\label{sec:sam}
We provide a study about using masks from Segment Anything (SAM) instead of the originally provided masks of the datasets. SAM masks are generated using a combination of text and flow-based conditioning. We estimate the first mask of the sequence using a text-based SAM approach\footnote{https://github.com/luca-medeiros/lang-segment-anything/tree/main}, and use this mask as the starting point. We then sample some points from within this mask, making sure that the points are not close to each other by sampling at regular intervals. Then, we use optical flow between the current image and the next image to find where these points will move. The moved points are then used as conditioning for SAM at the next timestep. Along with this, we also do text-based segmentation to obtain the same image masks. As some masks were not temporally consistent by either the flow or text-based method, we do a logical OR operation between both to make sure that all regions are covered.

Interestingly, the masks obtained for the synthetic cactus sequence using the described procedure with SAM are even slightly better than the ground truth masks, based on the experimental results shown in Tab.~\ref{tab:sam_comparison}. While in general, the reconstruction quality with SAM mask vs. GT mask is very similar, in this case, the SAM masks slightly outperform the GT masks. The reason for this can be seen in Fig.~\ref{fig:sam_comparison}. While the GT mask is not fully correct and includes some part of the occluding bookshelves in the bottom, the SAM mask correctly segments the object.


\begin{figure*}[ht]
\centering
\includegraphics[width=\textwidth]{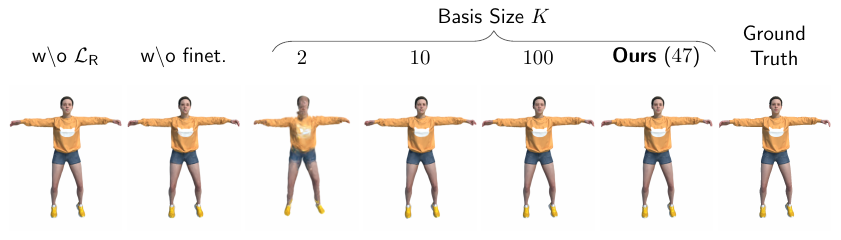}
\caption{\textbf{Qualitative ablation study on the D-NeRF dataset.} We show ablation on a novel view of the \textit{Jumping Jack} sequence with regards to different template optimization configurations and deformation basis size $K$.}
\label{fig:ablation-basis-vectors-d-nerf}
\end{figure*}
\begin{figure}[h]
\centering
\includegraphics[width=0.5\textwidth]{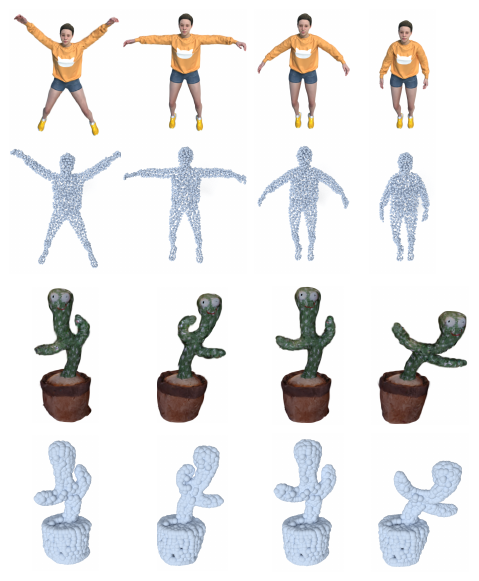}
\caption{\textbf{Optimized point templates.} \textbf{Top two rows:} Novel view renderings and point templates for the Jumping Jack sequence from the D-NeRF dataset. \textbf{Bottom two rows:} Novel view renderings and point templates for the Synthetic Cactus sequence from the Unbiased4D dataset. \vspace{-0.2cm}}
\label{fig:point_templates_cactus}
\end{figure}

\section{Additional Comparisons}
\label{sec:more-comparisons}
\subsection{High-resolution comparison}
\begin{figure}[h]
\centering
\includegraphics[width=0.5\textwidth]{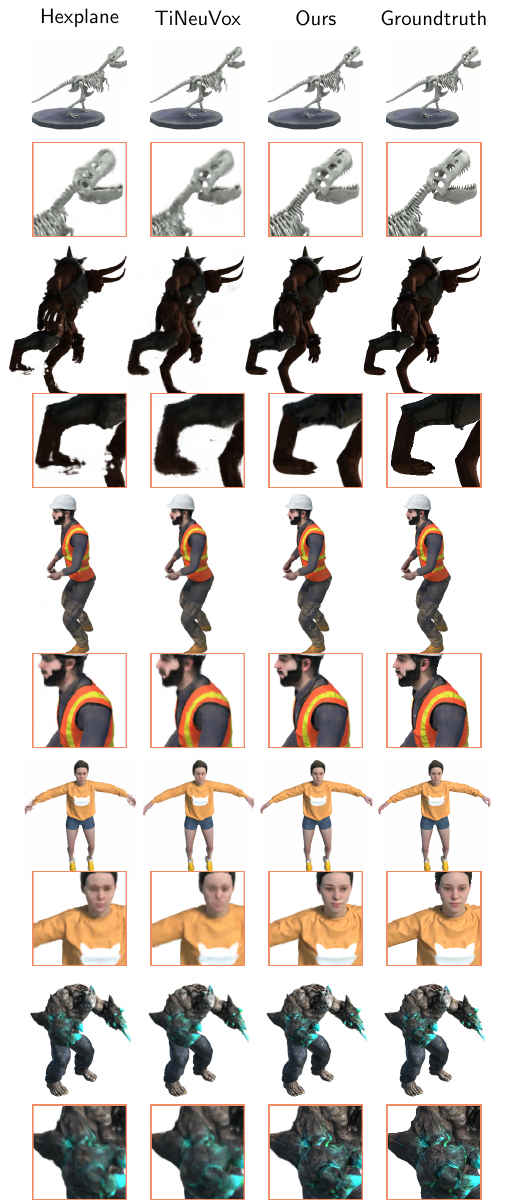}
\caption{\textbf{Qualitative comparison on 800 $\times$ 800 resolution for the D-NeRF dataset.} We show renderings of the \textit{T-Rex}, \textit{Hell Warrior}, \textit{Stand Up}, \textit{Jumping Jack} and \textit{Mutant} sequences.}
\label{fig:d-nerf_comparison_high_res}
\end{figure}

In addition to the comparisons on $400\times 400$ resulution in the main paper, we provide a D-NeRF comparison on $800 \times 800$ resolution, which is the full resolution of the dataset. We chose $400\times 400$ in the main paper, as it is the common evaluation setup in previous work. The quantitative results for $800 \times 800$ are shown in Tab.~\ref{tab:d-nerf_high_res}, while qualitative results are analyzed in Fig.~\ref{fig:d-nerf_comparison_high_res}. It can be seen that in high resolution, we outperform previous methods more clearly. This can be mostly attributed to the 3D Gaussian representation, which is capable of representing high frequency details well. These results are obtained with $N = 5000$.

\subsection{Static ground plane in D-NeRF dataset}
In the main paper, we chose the 5 scenes from D-NeRF which fit our single object setting. There are two additional scenes that only contain one object, the Lego and the T-Rex scene. However, these two contain ground planes, which make experiments on these two scenes challenging: the provided masks are dominated by the ground plane that covers most of the object when viewed from higher angles. In order to make a comparison anyway, we use the SAM mask procedure described in Sec.~\ref{sec:sam} to obtain a mask for the dynamic object on top of the ground plane only. Then, we apply our method to the object and use a static Gaussian splatting representation for the rest of the scene. Note that the static Gaussian representation can be easily modeled within our method by introducing an additional, static volume defined by 20 points on a cube around the scene. The results are given in Tab.~\ref{tab:d-nerf_high_res}. It can be seen that we obtain competitive results on these scenes as well. We provide both scenes qualitatively in our supplemental video.


\section{Optimized Coarse Point Models}
\label{sec:point-templates}

Fig.~\ref{fig:point_templates_cactus} shows the obtained point templates for two sequences containing challenging motion. Our templates exhibit temporal consistency and provide strong regularization for novel view renderings of the scene. These results are obtained with $N= 1500$.

\begin{figure}[t]
  \centering
  \includegraphics[width=\linewidth]{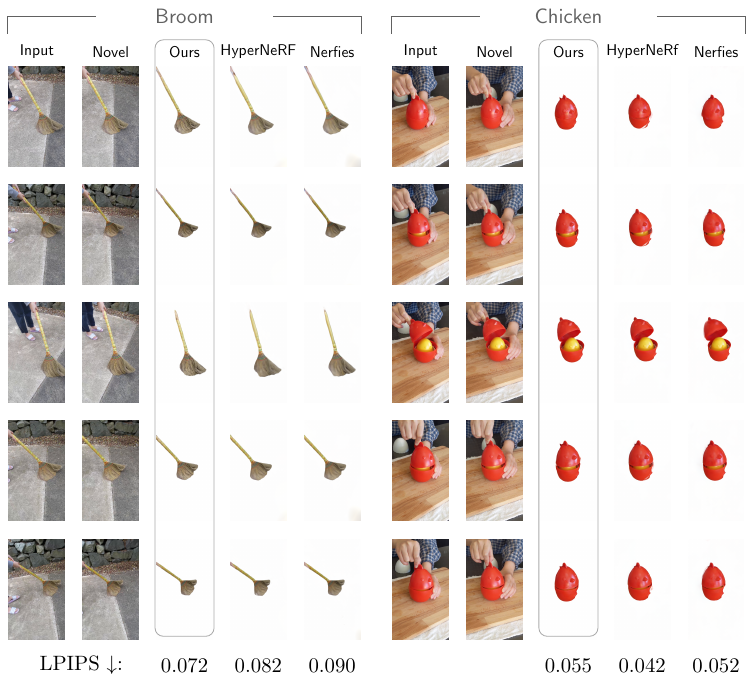}\vspace{-0.3cm}
   \caption{Additional comparison on sequences from HyperNeRF dataset.\vspace{-0.3cm}}
   \label{fig:hypernerf}
\end{figure}

\section{Additional Comparisons on Real Scenes}
\label{sec:Hypernerf-comparisons}
In addition to the provided comparisons we also provide additional examples of real objects from the HyperNeRF dataset in Fig.~\ref{fig:hypernerf} to show that our method works in more diverse scenarios. We have masked the background during training of the methods we are comparing against to ensure a fair comparison against our approach. We have used SAM \cite{kirillov2023segment} to obtain the masks of the foreground object.

\section{Interpolation of Basis Coefficients}
\label{sec:Basis_coeff_interp}
\begin{figure}[t]
  \centering
  \includegraphics[width=\linewidth]{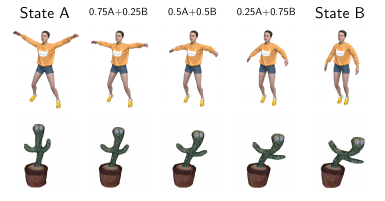}\vspace{-0.3cm}
   \caption{Interpolation of basis coefficients between two states.\vspace{-0.3cm}}
   \label{fig:interp_basis_coeff}
\end{figure}

We provide interpolations of basis coefficients in  Fig.~\ref{fig:interp_basis_coeff}. While interpolation is possible, \textit{linear} interpolation cannot lead to correct, \textit{non-linear} pose deformation (e.g. arms get shorter and longer). In the full method, correct deformation is modeled through non-linear change of coefficients, predicted by the MLP.

\section{Ablation Study on the D-NeRF dataset}
\label{sec:more-ablation-studies}

\subsection{Optimization and Deformation Basis Size}
In Tab.~\ref{tab:ablation_d-nerf}, we provide the ablation study on the Jumping Jack sequence from the D-NeRF dataset, similar to the ablation study on the Unbiased4D dataset in the main paper. Here, we find that 47 is the optimal basis size. Note that although the PSNR is higher with basis size 100, Fig.~\ref{fig:ablation-basis-vectors-d-nerf} shows that the details in the face are lost, while basis size 47 preserves details in the face, also indicated by the better SSIM and LPIPS score. This experiment is carried out with $N= 1500$.

\input{supplementary/tables/ablation_d-nerf}
\input{supplementary/tables/ablation_template_cactus}

\subsection{Coarse Point Model Size}
\begin{figure}[ht]
\centering
\includegraphics[width=0.5\textwidth]{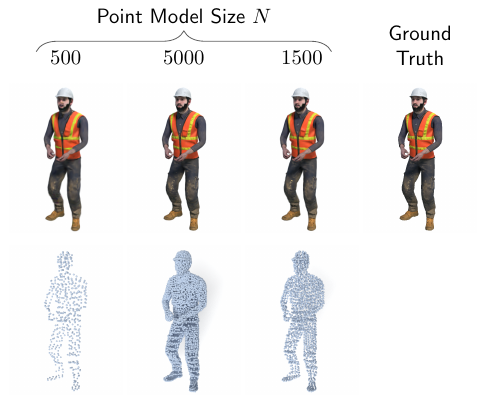}
\caption{\textbf{Qualitative results for different point model sizes.} \textbf{Top row:}  Novel views for configurations with different numbers of points $N$ for the \emph{Standup} scene of D-NeRF. \textbf{Bottom row:} Corresponding coarse point models.\vspace{-0.2cm}}
\label{fig:point_template_ablation}
\end{figure}

Last, we provide an ablation on number of points in our coarse point model. We show different configurations on the \emph{Standup} scene quantitatively in Tab.~\ref{tab:ablation_template} and qualitatively in Fig.~\ref{fig:point_template_ablation}. It can be seen that the results improve further with higher numbers of points. The results in the main paper are obtained with $N=5000$.
\input{supplementary/tables/ablation_template}

This is however not the case for synthetic cactus. We further found that for \textit{Synthetic Cactus} of the Unbiased4D dataset, 1500 points are a sweet spot. The quantitative results are shown in Tab.~\ref{tab:ablation_template_cactus}.

%% file: supplementary/tables/sam_comparison.tex
\begin{table}[t]
\small
\centering

\setlength{\tabcolsep}{3.85pt}

\begin{tabular}{l|ccc}
\toprule 
Method & PSNR$\uparrow$ & SSIM$\uparrow$ & LPIPS$\downarrow$   \\ 
\midrule
Ours with GT Masks & \textbf{22.445} & 0.90 & 0.099  \\ 
\rowcolor{rowhighlight}%
\textbf{Ours} with SAM masks & 22.348 & \textbf{0.905} & \textbf{0.095}\\
\bottomrule
\end{tabular}

\caption{\textbf{SAM vs. Ground Truth Mask} A comparison of masks on the synthetic cactus sequence from Unbiased4D dataset. In general we observe that our method works well with masks generated with SAM. In this specific case, it even slightly outperforms the results obtained with the ground truth mask.}
\label{tab:sam_comparison}
\vspace{-0.3cm}
\end{table}

%% file: supplementary/tables/d-nerf_high_res.tex
\begin{table*}[h]
\small
\centering
\scalebox{0.925}{
\begin{tabular}{l|ccc|ccc|ccc|ccc}
\toprule
& \multicolumn{3}{c|}{Jumping Jacks} & \multicolumn{3}{c|}{Hell Warrior} & \multicolumn{3}{c|}{Hook} & \multicolumn{3}{c}{Stand Up} \\ 
Method & PSNR$\uparrow$ & SSIM$\uparrow$ & LPIPS$\downarrow$ & PSNR$\uparrow$ & SSIM$\uparrow$ & LPIPS$\downarrow$ & PSNR$\uparrow$ & SSIM$\uparrow$ & LPIPS$\downarrow$ & PSNR$\uparrow$ & SSIM$\uparrow$ & LPIPS$\downarrow$ \\ 
\midrule
D-NeRF~\cite{cvpr_PumarolaCPM21} & 32.70 & 0.9779 & 0.0388 & 24.06 & 0.9440 & 0.0707 & 29.02 & 0.9595 & 0.0546& 33.13 & 0.9781 & 0.0355 \\ 
TiNeuVox ~\cite{TiNeuVox} & 33.49 & 0.9771 & 0.0408 & 27.10 & 0.9638 & 0.0768 & 30.61 & 0.9599 & 0.0592& 34.61 & 0.9797 & 0.0326 \\ 
HexPlane~\cite{cao2023hexplane} & 31.00 & 0.9723 & 0.0419 & 23.51 & 0.9479 & 0.0786 & 27.63 & 0.9490 & 0.0630& 32.84 & 0.9792 & 0.0308 \\
\rowcolor{rowhighlight}%
\textbf{Ours} & \textbf{33.97} & \textbf{0.9828} &	\textbf{0.0345}
 & \textbf{38.68} &	\textbf{0.9780} & \textbf{0.0537}
 & \textbf{33.39} & \textbf{0.9735} & \textbf{0.0460}&\textbf{38.20} & \textbf{0.9889} & \textbf{0.0257}
 \\

\toprule
& \multicolumn{3}{c|}{Mutant} & \multicolumn{3}{c|}{Lego} & \multicolumn{3}{c|}{T-Rex} & \multicolumn{3}{c}{Average}\\ 
Method & PSNR$\uparrow$ & SSIM$\uparrow$ & LPIPS$\downarrow$ & PSNR$\uparrow$ & SSIM$\uparrow$ & LPIPS$\downarrow$ & PSNR$\uparrow$ & SSIM$\uparrow$ & LPIPS$\downarrow$ & PSNR$\uparrow$ & SSIM$\uparrow$ & LPIPS$\downarrow$ \\ 
\midrule
D-NeRF~\cite{cvpr_PumarolaCPM21} 
& 30.31 & 0.9672 & 0.0392 
& 25.56 & \textbf{0.9363} & 0.0821
& 30.61 & 0.9671 & 0.0535 
& 29.34 & 0.9614 & 0.0534 \\ 
TiNeuVox ~\cite{TiNeuVox} 
& 31.87 & 0.9607 & 0.0474 
& \textbf{26.64} & 0.9258 & 0.0877
& 31.25 & 0.9666 & 0.0478 
& 30.8 & 0.9619 & 0.0560\\ 
HexPlane~\cite{cao2023hexplane} 
& 32.04 & 0.9670 & 0.0425 
& 24.81 & 0.9350 & \textbf{0.0621}
& 30.23 & 0.9699 & 0.0418 
& 28.87 & 0.9600 & 0.0515
 \\ 
\rowcolor{rowhighlight}%
\textbf{Ours} 
& \textbf{36.02} & \textbf{0.9840} & \textbf{0.0311}
& 24.63 & 0.9312 &	0.0716
& \textbf{32.10} &	\textbf{0.9818} & \textbf{0.0310}
& \textbf{33.86} & \textbf{0.9743} & \textbf{0.0419} \\

\bottomrule
\end{tabular}
}
\caption{\textbf{Novel view synthesis on the synthetic D-NeRF dataset.} We evaluate on the full 800 $\times$ 800 resolution and demonstrate superior performance compared to other methods. We also include two more scenes here, \textit{Lego} and \textit{T-Rex}, which include a static ground plane.}
\label{tab:d-nerf_high_res}
\vspace{-0.3cm}
\end{table*}

%% file: supplementary/tables/ablation_d-nerf.tex
\begin{table}[h]
\small
\centering

\begin{tabular}{l|ccc}
\toprule
Method & PSNR$\uparrow$ & SSIM$\uparrow$ & LPIPS$\downarrow$\\ 
\midrule
Ours w/o Rigidity Loss & 32.49 & 0.9770 & 0.0419 \\ 
Ours w/o Template Fine Tuning & 30.09 & 0.9660 & 0.0485 \\ 
Ours ($K=100$) & \textbf{33.94} & 0.9825 & 0.0355 \\
Ours ($K=10$) & 30.11 & 0.9697 & 0.0479 \\ 
Ours ($K=2$) & 24.29 &  0.9404 & 0.0769 \\ 
\rowcolor{rowhighlight}%
\textbf{Ours ($K=47$)} & 33.68 & \textbf{0.9844} & \textbf{0.0332} \\
\bottomrule
\end{tabular}

\caption{\textbf{Ablation study for template optimization and deformation basis size $K$.} We show performance for different configurations on the \textit{Jumping Jack} sequence of the D-NeRF dataset.}
\label{tab:ablation_d-nerf}
\vspace{-0.3cm}
\end{table}

%% file: supplementary/tables/ablation_template_cactus.tex
\begin{table}[t!]
\small
\centering

\begin{tabular}{l|ccc}
\toprule
Method & PSNR$\uparrow$ & SSIM$\uparrow$ & LPIPS$\downarrow$\\ 
\midrule
Ours ($N=500$) & 17.343 & 0.8611 & 0.1324 \\ 
Ours ($N=1500$) & \textbf{22.348} & \textbf{0.905} & \textbf{0.095} \\
Ours ($N=5000$) & 21.474 & 0.8946 & 0.1030 \\%
\bottomrule
\end{tabular}

\caption{\textbf{Ablation study on coarse point model size $N$.} Quantitative results for configurations with different numbers of points for the \emph{Synthetic Cactus} scene of Unbiased4D.}
\label{tab:ablation_template_cactus}
\vspace{-0.3cm}
\end{table}

%% file: supplementary/tables/ablation_template.tex
\begin{table}[h]
\small
\centering

\begin{tabular}{l|ccc}
\toprule
Method & PSNR$\uparrow$ & SSIM$\uparrow$ & LPIPS$\downarrow$\\ 
\midrule
Ours ($N=500$) & 33.90 & 0.9748 & 0.0435 \\ 
Ours ($N=1500$) & 36.14 & 0.9844 & 0.0332 \\
Ours ($N=5000$) & \textbf{38.20} & \textbf{0.9889} & \textbf{0.0257} \\
\rowcolor{rowhighlight}%
\bottomrule
\end{tabular}

\caption{\textbf{Ablation study on coarse point model size $N$.} Quantitative results for configurations with different numbers of points for the \emph{Standup} scene of D-NeRF.}
\label{tab:ablation_template}
\vspace{-0.3cm}
\end{table}